\documentclass{article}

\usepackage[nonatbib, preprint]{neurips_2021}  %

\usepackage[utf8]{inputenc} %
\usepackage[T1]{fontenc}    %
\usepackage{hyperref}       %
\usepackage{url}            %
\usepackage{booktabs}       %
\usepackage{amsfonts}       %
\usepackage{nicefrac}       %
\usepackage{microtype}      %
\usepackage{macros}

\title{Node Classification Meets Link Prediction on Knowledge Graphs}

\author{%
  Ralph Abboud, {\.I}smail {\.I}lkan Ceylan \\
  Department of Computer Science\\
  University of Oxford, UK\\
  \texttt{firstname.lastname@cs.ox.ac.uk} \\
}

\begin{document}

\maketitle
\begin{abstract}
Node classification and link prediction are widely studied in graph representation learning. While both transductive node classification and link prediction operate over a \emph{single} input graph, they have so far been studied separately. Node classification models take an input graph with node features and incomplete node labels, and implicitly assume that the graph is relationally complete, i.e., no edges are missing. By contrast, link prediction models are solely motivated by  relational incompleteness of the input graphs, and do not typically leverage node features or classes.  
We propose a unifying perspective and study the problems of (i) transductive node classification over \emph{incomplete graphs} and (ii) link prediction over graphs with \emph{node features}, introduce a new dataset for this setting, \datasetname, and conduct an extensive benchmarking study.
\end{abstract}

\section{Introduction}
Node classification is one of the most widely studied tasks in graph representation learning \cite{HamiltonBook}, which is categorized as transductive or inductive, depending on whether the model has access to all graph nodes during training \cite{HamiltonYL17}. We focus on the transductive setting, where the input is a \emph{single, fixed} graph with node features and incomplete node labels, and where the goal is to predict the missing node labels, while assuming access to \emph{all} nodes during training. Graph neural networks \cite{GilmerSRVD17,VelickovicICLR18,Kipf16} are the most prominent models for node classification tasks, and transductive node classification has been studied for various types of graphs, including heterogeneous, multi-relational graphs \cite{YunJKKK19}.  The implicit assumption in transductive node classification is that the input graph is \emph{relationally complete}, i.e., no edges are missing from the input graph.

We argue that the relational completeness assumption in node classification is unrealistic for most graph domains including heterogeneous, multi-relational graphs, and introduce the problem of node classification over \emph{incomplete} graphs. This task is more challenging than standard node classification, since, in this case, the node classification model needs to additionally deal with the incompleteness of the underlying graph, while predicting node labels, so as to maximally benefit from implicitly (and correctly) completing the relational knowledge. Notice how this naturally bridges node classification on multi-relational graphs with link prediction, which is the task of inferring missing edges from an input knowledge graph, by exploiting the existing information in the graph \cite{TransE-NIPS13,TuckER}.

This unifying perspective has more to offer than empowering node classification models with the capability of link prediction. Technically, we can view heterogeneous, multi-relational graphs as knowledge graphs enhanced with node features, and so another interesting question is whether the task of link prediction can be informed by node features and classes. This leads to yet another problem formulation investigated in this paper, namely, the problem of link prediction over knowledge graphs with \emph{node features} and \emph{classes}, to improve link prediction performance over knowledge graphs using this additional information.

All in all, the main contributions of this paper can be summarized as follows: 
\begin{itemize}[--,leftmargin=15pt]
    \item We propose a unifying perspective for node classification and link prediction, and study (i) node classification over \emph{incomplete} graphs, and (ii) link prediction using \emph{node features} and \emph{classes}.
    \item  We carefully construct a knowledge graph, \datasetname, to evaluate models on transductive node classification and link prediction, and show that it is well-suited for both tasks. 
    \item We propose a new framework, \framework, that extends any shallow embedding model X to process features, based on a multi-layer perceptron (MLP). We then show that \framework models are very competitive with GNN models on \datasetname for node classification, and some even \emph{outperform} these models as relational incompleteness is introduced. 
    \item  We investigate the special case of \emph{entity classification}, where we drop node features altogether, and observe that \framework models achieve superior performance against GNNs across all degrees of relational incompleteness. 
    \item We then focus on \emph{link prediction} on \datasetname, and show that \framework models significantly outperform their corresponding base models by additionally leveraging node features.
\end{itemize}

To our knowledge, our work is the first benchmarking study that combines node classification and link prediction on knowledge graphs, and paves the way for future research across the whole spectrum of node embedding models ranging from shallow to deep embedding models.  

\section{Node Classification and Link Prediction on Knowledge Graphs} 

\noindent\textbf{Knowledge graphs with features.} Consider a relational vocabulary, which consists of finite sets $\Ebf$, $\Cbf$, $\Rbf$ of \emph{entities}, \emph{classes}, and \emph{binary relations} respectively.
A \emph{unary fact} is of the form $c(e)$, where $c \in \Cbf$ is a class (or, a unary relation) and $e \in \Ebf$ is an entity.  A \emph{binary fact} is of the form $r(e_h, e_t)$, where $r \in \Rbf$ is a binary relation and $e_h,e_t \in \Ebf$ are entities. It is common to refer to $e_h$ as the \emph{head} entity, and $e_t$ as the \emph{tail} entity in a fact $r(e_h, e_t)$.
A \emph{knowledge graph (KG)} $G$ is a finite set of \emph{facts} defined over a \emph{relational vocabulary}\footnote{Note that most existing benchmark knowledge graphs do not explicitly encode unary facts. We nonetheless focus on this more general formulation, as it offers a unifying perspective.}.
Equivalently, a knowledge graph can be viewed as a \emph{labeled} graph, where nodes correspond to entities, node labels correspond to classes, and edge labels correspond to relations.
Moreover, we consider knowledge graphs potentially enriched with \emph{node features}, where for each entity $e \in \Ebf$, there exists a $k$-dimensional \emph{feature vector} $x \in \Rbb^k$ describing the entity. 
For simplicity, we denote the set of all node features as a matrix $\Xbf \in \Rbb^{|\Ebf|\times k}$. 

\noindent
\textbf{Problems.} The focus of this paper is the following problems defined over knowledge graphs.
Given a KG with features, where a subset of entities are labeled with classes, \emph{node classification} is the task of predicting classes for entities whose class is not already known. \emph{Entity classification} is a special case of node classification, where the input KG has no features~\cite{SchlichtkrullKB18}.
Given a KG, \emph{link prediction}, or \emph{knowledge graph completion~(KGC)}, is the task of accurately predicting missing (binary) facts.

Node classification and link prediction can be studied in the \emph{transductive}, or in the \emph{inductive} setting \cite{HamiltonYL17}. In the transductive setting, all entities are seen by a model during training, and therefore testing solely consists of predicting classes and links between known entities. In the inductive setting, new unseen entities could potentially appear during evaluation, and the model is expected to generalize and make predictions over these entities. The focus of this work is on the transductive setting. 

Motivated by the \emph{incomplete} nature of many real-world KGs, we investigate node and entity classification over KGs, where a fraction of the edges is missing from the input graph. This seemingly minor reformulation results in a substantially harder task, as it requires a model that can also deal with relational incompleteness in the underlying graph. We also investigate link prediction in a more general sense, in that we consider KGs with classes and features, allowing a link prediction model to be informed by features towards learning better representations leading to higher quality predictions.

\section{Related Work and Motivation} 
Graph neural networks are used for both node classification and link prediction. Additionally, there are a plethora of node embedding models targeting link prediction. We briefly review these models.

\noindent\textbf{Graph neural networks.} In recent years, graph neural networks \cite{Gori2005,Scarselli09} have become \emph{de facto} standard models for node and entity classification, and are also used for link prediction \cite{SchlichtkrullKB18}. We briefly introduce the framework of message passing neural networks (MPNNs) \cite{GilmerSRVD17} which encompasses several popular graph neural network architectures. 
In a MPNN, input graph nodes are each given vector representations, and these representations are updated through a series of message passing iterations. 
Formally, given a node $x$, its vector representation $v_{x,t}$ at iteration $t$, and its neighborhood $N(x)$, a message passing iteration can be written as:
\[
v_{x,t+1} = combine \Big(v_{x,t},aggregate\big(\{v_{y,t}|~y \in N(x)\}\big)\Big),
\]
where \emph{combine} and \emph{aggregate} are functions, and \emph{aggregate} is typically permutation-invariant.

Prominent MPNNs include graph convolutional networks~(GCN)~\cite{Kipf16}, and graph attention networks (GAT) \cite{VelickovicICLR18}. Indeed, GCN  and GAT achieved state-of-the-art results for node classification on citation graph benchmarks. However, these models cannot handle \emph{multi-relational} graph data, as they do not make any attribution to edge types. To address this limitation, relational GNNs, such as rGCNs \cite{SchlichtkrullKB18}, have been proposed, which perform aggregation based on relation types. Recently, these models have been extended to incorporate attention \cite{WangJSWYCY19} and transformer-based operations \cite{YunJKKK19}. 

GNNs have separately been used for link prediction \cite{KBGAT-ACL19, SchlichtkrullKB18, TeruDH20}, but these models cannot jointly tackle both tasks by design. Hence, GNNs for node classification are limited to aggregating over \emph{known edges} in the graph, without any attribution for potentially missing edges -- implicitly employing a relational completeness assumption.

\noindent\textbf{Knowledge graph embedding models.} Knowledge graph embedding (KGE) models represent the entities and relations of a KG using \emph{embeddings}, and  apply a \emph{scoring function} to these embeddings to compute likelihood scores for all possible KG facts. Then, these models train on all known KG facts (as well as negative sampled facts obtained by corrupting fact entities) to maximize their likelihood score in the embedding space. KGE models can broadly be classified into three main categories: translational models, bilinear models, and neural models. 

Translational models, such as TransE \cite{TransE-NIPS13} and RotatE \cite{RotatE-ICLR19}, represent entities as points in a low-dimensional space, and relations as translations in this space, such that the score of a binary fact corresponds to the distance between the translated head and its tail. This translational approach has also been generalized to spatio-translational models, such as BoxE \cite{BoxE-NeurIPS20}, where the correctness of a fact depends on its representation \emph{position} in the embedding space.
Bilinear models are based on tensor factorization: they represent entities as vectors, and relations as matrices, such that the bilinear product between entity and relation embeddings yields a plausibility score for a given fact. Bilinear models include RESCAL \cite{RESCAL-ICML11}, ComplEx \cite{ComplEx-ICML16} and TuckER \cite{TuckER}. Finally, neural models  \cite{ConvE-AAAI18,EMLP-NIPS13} use a neural architecture to perform scoring over KG embeddings. 

KGE models are widely applied for link prediction, and achieve state-of-the-art results on  standard KG benchmarks.  However, most existing models operate exclusively on binary facts, with no attribution to unary facts. Therefore, these models can only predict entity classes by modelling them also as binary relations, which is not very natural. 
There are only few models that address this limitation. BoxE~\cite{BoxE-NeurIPS20} can explicitly handle class information, by viewing classes as boxes in the space.
TransC~\cite{LvHLL18} extends TransE with hyper-spheres to explicitly represent classes, but has limited expressive power, and is not competitive with state-of-the-art models. 
Most importantly, however, none of these models and benchmarks consider \emph{node features} as part of KGs for link prediction. 

Overall, we offer a unified view for node classification and link prediction to address these limitations. We propose a framework to tackle these problems, and introduce a dataset for conducting a detailed benchmarking. Our analysis reveals the practical relevance of the above-mentioned limitations. 

\section{Embedding Models for Node Classification and Link Prediction}
\label{sec:mlpx}

In this section, we propose the \framework framework, which extends embedding models to leverage node features, jointly train on transductive node classification and link prediction, and subsequently solve both problems in a unified fashion. Fundamentally, \framework introduces an MLP to process features, and combines MLP output with node embeddings of X using a weighted summation. Furthermore, \framework modifies negative sampling to better match class structures and obtain a better inductive bias during training. We provide a general overview of \framework, refer the reader to the appendix of this paper for model-specific instantiations (e.g., MLP-TransE) and more detailed explanations.

\noindent\textbf{\framework models.} In a knowledge graph embedding model, every entity $e_i \in \Ebf $ is typically represented with a $d-$dimensional vector $\bm{e_i} \in S^d$, in a space $S$, e.g., usually the real or complex space. This vector is learnable, and trained independently of node features. Therefore, we introduce an MLP $f: \Rbb^k \to S^d$, which maps input features into the latent embedding space, and combine its output with the learnable parameter using weighted summation\footnote{Note that some models, such as BoxE, define multiple representations per entity. In this case, we introduce as many MLPs as there are representations and parallelize our framework accordingly.}. More formally, for each entity $e_i$, a representation is defined as
    $\bm{e_i'} = \lambda \bm{e_i} + f(\bm{x_i})$,
where $\lambda \in \Rbb^{+}$ is a tunable hyper-parameter and  $\bm{x_i} \in \Rbb^k$ is the feature vector of $e_i$.

Intuitively,  \framework integrates feature processing (via MLPs) and structure encoding (via embeddings). More specifically,  MLPs process features in a \emph{structure-agnostic} fashion, and provide latent representations for feature information, whereas embeddings capture structural and relational information about a node, based on its role in the knowledge graph. Both components contribute, via the weighted summation, to an overall entity representation which holistically captures node properties. 

\noindent\textbf{Relational inductive bias.} \framework 
maintains the relational inductive bias of shallow embedding models. In particular, \framework models, via their embeddings, capture semantic structure and relation properties, similarly to their base models, and hence can also \emph{predict} unknown facts, based on entity and relation representations. Furthermore, feature processing does not introduce additional modeling assumptions, as MLPs do not rely on known edges or other nodes. By contrast, GNNs explicitly rely on the known structure, and process features using a relational completeness assumption. 

\noindent\textbf{Expressive power.} \framework maintains the \emph{expressive power} of its base models: if X is \emph{fully expressive}, i.e., for any disjoint sets of true and false
facts, there exists an X configuration that accurately classifies these facts, then \framework is also fully expressive. Furthermore, any set of facts \emph{not} captured by X cannot be reliably captured using \framework. This intuitively holds as the representations in \framework are in the same space as X: MLPs are universal \cite{cybenko1989}, and thus can map features to arbitrary values in the latent space. However, they cannot discriminate between structurally distinct nodes with identical features. Hence, \framework only maintains the expressiveness provided by learned embeddings.

In addition to processing features, \framework introduces the following changes in fact representation and negative sampling to enable embedding models to better capture class structure. 

\noindent\textbf{Handling classes.} 
For embedding models that \emph{cannot} represent classes (such as TransE), \framework performs \emph{binarization}, where each unary fact $c(e_i)$ is viewed as a binary fact $r_c(e_i, \gamma_c)$, where $r_c$ is a binary relation, and $\gamma_c$ is an auxiliary entity, both of which are specific for the class $c$.

\noindent\textbf{Enhanced negative sampling.} In node classification, a standard assumption is that classes are \emph{mutually exclusive}. Based on this assumption, we can define a better negative sampling process for KGE models, where the idea is to sample \emph{classes}, rather than entities, for unary facts. That is, rather than sampling, for $c(e_i)$, a negative fact $c(e_i'), i' \neq i$, as in KGE models, we sample $c'(e_i), c' \neq c$. The latter is guaranteed to result in a false fact, whereas the former cannot guarantee this.

\section{\datasetname: A Knowledge Graph with Node Features}
\label{sec:dataset}

In the literature, node classification models have primarily been evaluated on citation networks such as Cora, Citeseer, and PubMed \cite{SenNBGGE08,GNNPitfalls}, whereas link prediction models have been evaluated using KG benchmarks, such as FB15k-237 \cite{FB15k237TC} and WN18RR \cite{ConvE-AAAI18}. These benchmarks are not suitable for our joint perspective for various reasons, e.g., citation networks are single-relational (i.e., no relation types present), while link prediction benchmarks do not incorporate node classes or features. 

Rather recently, new multi-relational node classification benchmarks based on citation graphs, such as OGBN-MAG \cite{OGB-NeurIPS2020}, have been introduced. However, these benchmarks are limited in relation type information, as their relation types follow  directly from the classes of nodes they connect, e.g., paper-paper edges are ``cites'' relations. While some of these benchmarks are large, they have a very small number of distinct relations, e.g., OGBN-MAG only has 4 relation types. Therefore, neither link prediction nor node classification currently presents a holistic benchmark allowing a joint study of node classification with incompleteness, as well as link prediction with classes and node features.

Therefore, we propose a new dataset, \datasetname. \datasetname is a subset of the YAGO4 \cite{TanonWS20} English Wikipedia knowledge base, where the node classification target is to predict the \emph{alma mater} of individuals. %
Classes are produced using the $\mathsf{alumniOf}$ relation in YAGO4, and are \emph{mutually exclusive}, as \datasetname only includes entities with a \emph{unique} alma mater.

Class selection is based on two criteria. First, all classes must appear 700 or more times (in the tail of $\mathsf{alumniOf}$) in YAGO4, and thus be sufficiently represented. Second, classes must correspond to North American institutions to minimize homophily. Indeed, attending other institutions strongly correlates with sharing their home nationality, and thus we omit these to make the task more challenging. These criteria lead a final set of 24 classes, which we use to build \datasetname as follows:
\begin{enumerate}[leftmargin=18pt]
        \item For every class, we select the 700 nodes with the highest degrees in YAGO4, yielding a total of 16,800 target classification nodes evenly balanced between all 24 classes.
        \item We introduce all nodes connected to any of the 16,800 nodes by an edge. We exclude literal relations such as code strings, latitude, longitude, etc., and also exclude \emph{all} nodes appearing as the tail of $\mathsf{alumniOf}$ (even those that are not our target classes), to prevent solving the task by simply identifying the relevant institution neighbor. 
        \item Finally, we only keep facts with relations appearing 10 or more times among all previously selected edges. This yields 52,714 nodes, and 121,857 edges spanning 29 relations. 
\end{enumerate}
To produce node features, we scrape English Wikipedia to obtain the introduction text from entity pages. When this page does not exist ($\sim 0.2\%$ of nodes), we instead use its WikiData summary. Then, we convert text to feature vectors by mapping words to 300-dimensional GloVe embeddings \cite{PenningtonSM14} and taking their mean. Finally, we discard all entities without a description or GloVe embedding, and this yields a final dataset of 52,678 nodes and 121,836 edges. The overall \datasetname statistics are shown in \Cref{tab:dataset_stats}, and the detailed breakdown of relations and classes is provided in the appendix. 

\begin{table}[t]
    \centering
    \caption{Base statistics of the \datasetname dataset. Subsets \datasetname-90\% and \datasetname-80\% have analogous statistics, but only include 109,652 and 97,468 edges, respectively.}
    \label{tab:dataset_stats}
    \begin{tabular}{lcccccccc}
        \toprule
         \textbf{Dataset} &  Nodes & Edges & Classes & Relations & \makecell{Training \\ Labels} & \makecell{Validation \\ Labels} & \makecell{Testing \\ Labels} \\ %
         \midrule
         \datasetname & 52678 & 121836 & 24 & 29 & 10080 & 5040 & 1680\\
         \bottomrule
    \end{tabular}
\end{table}
In addition to \datasetname, we provide two carefully crafted splits, \datasetname-90\% and \datasetname-80\%, where we omit 10\% and 20\% of existing edges from training, respectively. These splits are made such that dropped edges do not produce any new isolated nodes. We make this choice to minimize reliance on pure feature-based computation, and to provide edge information for all nodes, to be used explicitly by GNNs, and implicitly by \framework models.  

All in all, \datasetname provides features and relational structure to support node classification, such that each helps solve distinct dataset subsets. Furthermore, it presents a non-trivial and rich relational structure, which cannot be determined from node features or classes. Finally, it is challenging for both link prediction and node classification, relative to their respective baseline models.

\section{Experimental Evaluation}
\label{sec:exp_ev}

In this section, we use \datasetname to evaluate \framework models, as well as state-of-the-art GNNs and baseline models, on the node classification task with edge incompleteness, as well as link prediction performance with/without node features and classes. 

\subsection{Node Classification with Incomplete Edges} 
\label{ssec:nc_inc}
\noindent\textbf{Experimental setup.}
 In this experiment, we train three \framework models, namely MLP-TransE, MLP-RotatE, and MLP-BoxE, on \datasetname, as well as on its incomplete splits \datasetname-90\% and \datasetname-80\%. We train using both class and edge information, and evaluate using the \emph{unary facts} in the \datasetname validation set. We additionally replicate this experiment in the \emph{entity classification} setting, where we discard node features and use the original embedding models TransE, RotatE, and BoxE, only keeping the refined negative sampling of \framework. In all experiments, we train \framework models using both negative sampling (NS) and cross-entropy (CE) loss, and report and discuss results for both losses.  
Finally, we compare against the following baseline models: 

\begin{itemize}[$\bullet$,leftmargin=15pt]
    \item \textbf{Label Propagation (LP)} \cite{zhu2002learning}: An iterative algorithm that exclusively uses node labels, i.e., classes, and edges (without  relations) to make class predictions. LP is agnostic to node features.  
    \item \textbf{Multi-Layer Perceptron (MLP)}: An MLP with two hidden layers of size 512. Trains exclusively on node features, and has no access to edge information.
    \item \textbf{rGCN} \cite{SchlichtkrullKB18}: A 2-layer relational GCN that aggregates neighborhood features based on edge type. We use rGCN as a baseline for heterogeneous GNNs. rGCN is regularized to use 10 basis vectors.
    \item \textbf{GAT} \cite{VelickovicICLR18}: A 2-layer GAT, where attention is used to aggregate neighborhood features. GAT does not explicitly use edge types, but performs strongly in practice on heterogeneous graphs, often outperforming heterogeneous GNN models \cite{YunJKKK19}. Hence, we include this model in our analysis.
\end{itemize}
Note that these baselines are dedicated to node and entity classification, and that GNNs are state-of-the-art models for these tasks. However, these baselines, unlike \framework models,  \emph{do not} additionally perform link prediction. In particular, GNNs only predict node classes, and can only be used for link prediction with a \emph{distinct} setup, where they train using a dedicated scoring function, e.g., DistMult \cite{SchlichtkrullKB18}, and maintain their relational completeness assumption. Hence, comparing between GNNs and \framework model performance trends will offer insights as to the impact of joint link prediction capacity on node classification performance under edge incompleteness.  

We use a fixed dimensionality of $d=128$ across all experiments for \framework models, rGCN, and GAT, and supplement GNN models with node embeddings in the entity classification setting, resulting in the models rGCN$^+$ and GAT$^+$, respectively. Further details about hyper-parameter setup and experimental protocol for all experiments can be found in the appendix.

\begin{table}[t]
    \centering
    \caption{Node and entity classification results (validation accuracy) for \framework models, using both NS and CE loss, and benchmark models on \datasetname and its incomplete splits. The top three results are reported in bold, with opacity increasing as rank improves.}
    \label{tab:featured_results}
    \begin{tabular}{l@{\hspace{2em}}c@{\hspace{2em}}c@{\hspace{2em}}c@{\hspace{2em}}l@{\hspace{2em}}c@{\hspace{2em}}c@{\hspace{2em}}c}
        \toprule
        \multicolumn{4}{c}{\textbf{Node Classification}} & \multicolumn{4}{c}{\textbf{Entity Classification}}\\
         \cmidrule(r){1-4}
         \cmidrule(r){5-8}
         Model &  80\% & 90\% & 100\% & Model &  80\% & 90\% & 100\%\\
         \cmidrule(r){2-2}
         \cmidrule(r){3-3}
         \cmidrule(r){4-4}
         \cmidrule(r){6-6}
         \cmidrule(r){7-7}
         \cmidrule(r){8-8}
         LP & 28.0 & 30.2 & 32.2 & LP & 28.0 & 30.2 & 32.2\\ 
         MLP & 35.5 & 35.5 & 35.5 & MLP & - & - & - \\
         \midrule
         rGCN & 39.0 & 39.7 &\first{42.4} & rGCN$^+$ & 24.8 & 27.1 & 30.5\\
         GAT & \third{40.0} & \second{41.2} &\first{42.4} & GAT$^+$ & 22.4 & 24.8 & 26.0\\ 
         \midrule
         MLP-TransE(NS) & 38.4 & 39.2 & 39.8 & TransE(NS) & \first{29.8} & \first{32.2} & \first{34.0} \\   
         MLP-RotatE(NS) & 37.2 & 38.5 & 39.1 & RotatE(NS) & \first{29.8} & \second{32.0} & \second{33.7}  \\ 
         MLP-BoxE(NS) & 38.8 & 39.3 & 40.4 & BoxE(NS) & \third{29.3} & \third{31.3} &  \third{33.4}\\ 
         MLP-TransE(CE) & \second{40.3} & \third{41.0} & 41.9 & TransE(CE) & 25.5 & 27.8 & 30.1 \\   
         MLP-RotatE(CE) & \third{40.0} & 40.7 & 41.4 & RotatE(CE) & 24.5 & 26.6 & 29.5  \\ 
         MLP-BoxE(CE) & \first{40.5} & \textbf{41.4} & \third{42.1} & BoxE(CE) & 26.5 & 29.3 & 31.0 \\ 
         \bottomrule
    \end{tabular}
\end{table}

 \noindent\textbf{Results.} The results for both node and entity classification with incomplete edges are shown in \Cref{tab:featured_results}. First, we note that overall results confirm the difficulty and balance  of \datasetname. In particular, feature-based accuracy (MLP, 35.5\%) and edge-based accuracy (LP, 32\% and EC results for \framework models, $\sim 33.5\%$) are both substantial, highlighting the importance of both features and edge structure. Furthermore, GNNs, which use both edge and feature information, achieve substantially higher scores (42.4\%), which confirms that the contributions of features and relational structure do not fully overlap. Finally, all models fail to achieve even 50\% accuracy on \datasetname,  highlighting the difficulty of the dataset. 
 
 In the node classification setting, rGCN and GAT achieve the strongest performance on the full \datasetname, with MLP-BoxE(CE) slightly behind, and MLP-TransE(CE) and MLP-RotatE(CE) performing competitively. This matches expectations, given the strong performance of GNNs and the relative completeness of the full \datasetname. However, as edges are dropped, the trend reverses: rGCN and GAT fall significantly, while \framework models are more robust. In fact, MLP-BoxE(CE) and MLP-TransE(CE) ultimately beat both GNNs on \datasetname-80\%. This suggests that the link prediction capability of \framework models alleviates the loss of dropped edges.  

\begin{wrapfigure}[25]{r}{0.47\textwidth}
    \centering
    \begin{tikzpicture}[scale=0.8, every node/.style={transform shape}]
\definecolor{EntityCol}{rgb}{1, 1, 1}
\definecolor{featCol1}{rgb}{0.38,0.62,0.92}
\definecolor{featCol2}{rgb}{0.38,0.82,0.12}
\definecolor{completionCol}{rgb}{0.8,0.2,0.0}
\tikzstyle{entity} = [text width=1.6em, text centered, circle, draw,inner sep=1pt]
\tikzstyle{feature} = [rectangle, draw opacity=0.3, draw, align=center]
\tikzstyle{shortened} = [shorten <= 7pt, shorten >= 7pt]
\tikzstyle{scoreText} = [font=\small\itshape, opacity=0.8]
\node[entity](ED){};
\node[entity,below left=1.95cm and 3.15cm of ED](US){};
\node[entity, below right=1.8cm and 3.15cm of ED](SO){};
\node[entity, below right =3.6 and 0.3 cm of ED](CB){};
\node[above=0.1cm of ED, align=center]{Edmund\\Dollard};
\node[below=0.1cm of US, align=center]{United\\States};
\node[below=0.1cm of SO, align=center]{Syracuse\\Orange};
\node[below=0.1cm of CB, align=center]{Coach\\(Basketball)};

\draw[line width=0.6mm,->,shortened] (ED) -- (US) node[midway, sloped, above]{nationality};
\draw[line width=0.6mm,->,shortened, scoreText] (ED) -- (US) node[midway, sloped, below]{4.33};

\draw[line width=0.9mm,->,shortened, dashed, completionCol] (ED) -- (SO) node[midway, sloped, above]{memberOf};
\draw[line width=0.9mm,->,shortened, dashed, completionCol, scoreText] (ED) -- (SO) node[midway, sloped, below]{3.99};

\draw[line width=0.35mm,->,dashed, shortened, completionCol] (ED) -- (CB) node[midway, sloped, above]{hasOccupation};
\draw[line width=0.35mm,->, dashed, shortened, completionCol, scoreText] (ED) -- (CB) node[midway, sloped, below]{5.19};

\begin{customlegend}[legend cell align=left, %
legend entries={ %
Training edge, Predicted edge},
legend style={at={(-0.4,-4)},font=\footnotesize}] %
    \addlegendimage{->, line width=0.3mm}
    \addlegendimage{dashed, ->, line width=0.3mm, completionCol}
\end{customlegend}
\end{tikzpicture}
    \caption{Visualization of MLP-BoxE completion on a sub-graph of \datasetname. Based on the features of  ``Edmund Dollard'' and KG structure, MLP-BoxE predicts his membership of Syracuse Orange, and also strongly predicts his occupation as a basketball coach. Both edge completions significantly simplify the class prediction of Syracuse University.}
    \label{fig:example}
\end{wrapfigure}
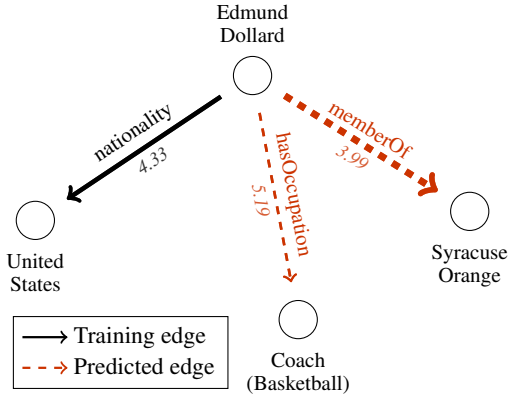

To illustrate this behavior, we perform a case study on a sub-graph from \datasetname, shown in \Cref{fig:example}, using the best-performing model, MLP-BoxE. In this sub-graph, entity ``Edmund Dollard'' appears with three neighbors, but only one  neighbor, the ``United States'' nationality, survives in \datasetname-80\%. Clearly, this neighbor is not sufficient to classify the entity, and therefore GNNs lose key information. 
By contrast, MLP-BoxE successfully completes these missing edges. 
In fact, it predicts the ``memberOf'' edge with a lower score (3.99) (lower implies higher confidence in this context) than the ``nationality'' edge it has trained on (4.33), and also predicts the ``hasOccupation'' edge with a comparable score (5.19). Hence, MLP-BoxE can leverage the predicted edges to classify ``Edmund Dollard'' as an alumnus of Syracuse University. 

This paints a subtle but informative picture regarding the viability of transductive node classification with GNNs. With more complete edge information, both rGCN and GAT perform high-quality aggregation and make good predictions without suffering from their lack of link prediction capacity. %
However, as data becomes incomplete, aggregation becomes noisier, thereby compromising prediction quality. This phenomenon can be observed on \datasetname. There, GAT and rGCN perform strongly, but then suffer significantly with incompleteness. In fact, rGCN falls much more than GAT, and this can also be explained by aggregation: as GAT aggregates using attention, it is less susceptible to noise, as it can ``ignore'' noisy neighbors. However, rGCN, by construction, considers all nodes, however irrelevant, and aggregates uniformly across relations. Hence, GAT is more resilient to dropped edges than rGCN, but in return is heavily reliant on node features. We  confirm this reliance with an additional experiment using bag-of-words features in the appendix. %

\framework models fall behind of GNNs with complete edges, which is sensible, as link prediction will not produce missing edges, and will essentially distract from node classification.
Yet, with incomplete data, these models can identify potentially critical missing edges and recover information, thereby enabling stronger node classification performance. Note, however, that this benefit is intrinsically tied to the quality of link prediction. Indeed, were a model to jointly make poor link predictions, then this would result in noisy, and even wrong, neighborhoods, which ultimately hurts node classification and makes performance deteriorate faster with incompleteness. Hence, the strong performance of \framework models suggests that these models are robust not simply because they \emph{can} perform link prediction, but rather because they do this \emph{with high quality}, as suggested by our case study.     

For entity classification, \framework models trained with negative sampling loss significantly outperform all other models, whereas GNNs, even with embeddings, fail to surpass the LP baseline. This suggests that link prediction capacity and relational inductive bias of embedding models offers a strong advantage, which, for entity classification, cannot be compensated via node features. Interestingly, negative sampling loss enables better performance for entity classification, while cross-entropy is better for node classification. This can be explained by the assumptions made by the respective losses. 
With negative sampling loss, the margin hyper-parameter sets \emph{absolute}, \emph{uniform} ranges for fact scores, such that  \emph{all} positive facts score slightly lower than the margin, and negative facts score slightly higher. Therefore, margin sets a score range for \emph{all} facts in the data, and thus provides additional information. By contrast, cross-entropy optimizes the \emph{relative} difference between positive and negative fact scores, \emph{without} setting an absolute score range, instead allowing scores to be \emph{data-driven}. Hence, each loss has properties that better suit different tasks. For entity classification, data is uniform (due to lack of distinguishing features), and thus negative sampling loss aligns better with the input, %
whereas cross-entropy is likelier to overfit, as it learns a more general score mapping from this data. However, for node classification, node features provide information that substantially helps cross-entropy learn \emph{data-driven} and \emph{variable} scores, whereas the uniformity of negative sampling loss score ranges does not align with the diversity of input features. 

Finally, we note that enhanced negative sampling is essential for achieving strong results with \framework models. In fact, without this sampling, \framework model performance for node classification drops around 2\%. This confirms the insights discussed in \Cref{sec:mlpx}, and highlights the importance of selecting higher-quality negative examples to enforce mutual exclusion between classes. %

\subsection{Link Prediction with Features and Classes}

 \noindent\textbf{Experimental setup.} We now evaluate the link prediction quality of \framework models on \datasetname-80\%, using the remaining edges dropped from \datasetname. As in the earlier experiment, we consider configurations with and without \emph{node features}, and train all \framework models using both negative sampling and cross-entropy loss. However, we additionally study configurations with and without \emph{node classes}. Therefore, we study four different configurations, relative to the use/omission of node features and classes. Finally, we evaluate using standard metrics, namely mean rank (MR), mean reciprocal rank (MRR), and Hits@10 (H@10) \cite{TransE-NIPS13}. 

\begin{table}[t]
    \centering
    \caption{Link prediction results for \framework models and their featureless counterparts on \datasetname-80\%, using both NS and CE loss. Results are categorized per feature/class configuration and, for each setup, the top result is reported in bold.}  %
    \label{tab:lp_results}
    \begin{tabular}{l@{\hspace{2em}}c@{\hspace{2em}}c@{\hspace{2em}}c@{\hspace{2em}}c@{\hspace{2em}}c@{\hspace{2em}}c}
    \toprule
    \cmidrule(r){2-7}
    \multirow{2}{*}{\textbf{Model}}& \multicolumn{3}{c}{Classes} & \multicolumn{3}{c}{No Classes}\\
    \cmidrule(r){2-4}
    \cmidrule(r){5-7}
     & MR & MRR & H@10 & MR & MRR & H@10\\
     \midrule
    TransE(NS) & \textbf{1544} & 0.265 & 0.382 & 1843 & 0.265 & \textbf{0.376}  \\  %
    RotatE(NS) & 1655 & \textbf{0.298} & \textbf{0.386} & 1806 & \textbf{0.285} & 0.375 \\ 
    BoxE(NS) & 1768 & 0.250 & 0.344 & \textbf{1780} & 0.245 & 0.341\\ 
    \midrule
    TransE(CE) & 2063 & 0.255 & 0.346 & 2162 & 0.246 & 0.340  \\
    RotatE(CE) & 2934 & 0.236 & 0.335 & 3673 & 0.230 & 0.317 \\
    BoxE(CE) & 2170 & 0.288 & 0.369 & 2298 & \textbf{0.285} & 0.366\\ 
    \midrule
    \midrule
    MLP-TransE(NS) & 930 & 0.285 & 0.407 & 988 & 0.275 & 0.402 \\
    MLP-RotatE(NS) & 898 & \textbf{0.316} & \textbf{0.417} & 932 & 0.314 & 0.414 \\ 
    MLP-BoxE (NS) & 1025 & 0.257 & 0.374 & 1077 & 0.256 & 0.368\\  %
    \midrule
        MLP-TransE(CE) & \first{747} & 0.242 & 0.378 & \textbf{781} & 0.235 & 0.370 \\
    MLP-RotatE(CE) & 753 & 0.291 & 0.416 & 816 & 0.288 & \textbf{0.416} \\
    MLP-BoxE(CE) & 883 & \first{0.316} & 0.415 & 947 & \textbf{0.315} & 0.413\\
    \bottomrule
    \end{tabular}
\end{table}

 \noindent\textbf{Results.} The results for all four link prediction configurations across all \framework models with both NS and CE loss are shown in \Cref{tab:lp_results}. Across all models and loss functions, features yield substantial improvements, more than halving the best MR and improving MRR by over 0.02, both with and without classes. Classes also slightly improve performance, and more so in the featureless setting, where, on average, MR improves by 200 (negative sampling loss), and Hits@10 improves by 0.007. 

These results align with our expectations. Features, useful for node classification, also support link prediction, as feature pairs produce rich context. 
In particular, knowing the features of two entities helps better establish similarities and discrepancies between them, which yields better entity representations. Hence, features provide a dual advantage: they simplify node classification by providing additional information, and also indirectly improve class predictions by enabling better link prediction, and subsequently improving node representations. Therefore, features are generally valuable for knowledge bases, and should be used to supplement existing KGC models and benchmarks when available, as can already be seen with recent multi-modal knowledge graphs \cite{LiuLGNOR19}.

With features and classes, all models perform strongly with CE loss, and achieve good mean rank (\~top 1.5-1.8\% of all possible entities) and Hits@10. This aligns with the earlier node classification results, and confirms our intuition regarding link prediction quality. %
In fact, we also observe the opposite phenomenon: RotatE(CE) performs poorly for entity classification (cf. \Cref{tab:featured_results}) and responds worse to incompleteness than its counterparts, and this indeed aligns with poorer link prediction results for RotatE(CE) in the configuration with classes.

We observe an interesting contrast in model behavior relative to the choice of loss function. Indeed, MLP-TransE and MLP-RotatE perform better with negative sampling loss in configurations without node features, analogously to  entity classification. However, with features, cross-entropy significantly improves MR but \emph{worsens} MRR, despite it substantially improving node classification performance for both models. This may initially appear surprising, but it can in fact be traced back to a combination of improved node classification with the limited expressiveness of MLP-TransE and MLP-RotatE: As cross-entropy improves node classification, unary facts on average receive lower scores. However, achieving lower unary fact scores for MLP-TransE and MLP-RotatE imposes, by construction, that entities belonging to a same class map closer to a same point. %
To see this, observe that fitting the binarized facts $r_c(a, \gamma_c)$  and $r_c(b, \gamma_c)$ in both models implies $a=b$ as both fact scores tend to zero. Hence, as cross-entropy improves node classification, entities in both models are \emph{more clustered}. This allows for better \emph{general} fact prediction, as entities are semantically better grouped, leading to \emph{better} MR. Yet, this reduces both models' ability to optimize individual facts, and therefore worsens MRR. Hence, the interplay between node classification and link prediction takes on an added dimension in these models, as their limited expressiveness enforces clustering with improved node classification, and ultimately makes their link predictions less fine-grained.

Interestingly, MLP-BoxE does not have the same behavior as MLP-TransE and MLP-RotatE, and in fact achieves its best performance with cross-entropy across all configurations. This can also be attributed to model properties. In fact, MLP-BoxE is fully expressive, and thus does not exhibit the clustering behavior described earlier. Furthermore, cross-entropy is better suited to the BoxE scoring mechanism than negative sampling loss. Indeed, 
BoxE assigns scores at every arity position and \emph{sums these up} to yield an overall fact score. Hence, the range of binary fact scores is much higher and larger than that of unary facts by design. Therefore, negative sampling loss, which sets absolute score ranges for \emph{all} positive facts, irrespective of arity, invariably creates competing objectives between unary and binary facts, leading to sub-optimal training. %
By contrast, cross-entropy is \emph{agnostic} to score values, and considers the relative difference between scores. Therefore, it provides a better optimization for BoxE, yielding better results. 

All in all, these results suggest that \framework models achieve strong performance on node classification and link prediction, despite their generality, and highlight that this unifying perspective leads to substantial benefits in terms of robustness and performance improvements.

\section{Summary, Discussions, and Outlook}
We studied (i) transductive node classification over \emph{incomplete graphs} and (ii) link prediction over graphs with \emph{node features}, unifying node classification and link prediction. 
We proposed a new framework, \framework, and dataset, \datasetname, and conducted an extensive benchmarking study.

Our findings %
highlight the importance of joint link prediction to tackle incompleteness, and raise a more general question about the best assumptions when solving problems over graphs. 
Specifically, GNNs only aggregate information between known neighbors, and hence implicitly adopt the \emph{closed world assumption (CWA)}~\cite{Reiter}. By contrast, \framework models adopt, and benefit from, the \emph{open world assumption (OWA)}. %
This is a very fundamental difference that motivates exploring different approaches more conceptually connected with embedding techniques, such as node2vec \cite{GroverL16} and metapath2vec \cite{DongCS17}, and proposing new, targeted, benchmarks within graph representation learning. 

\datasetname provides real-world features, rich relational structure, and balanced, representative classes, and therefore is a strong benchmark not only for our studied tasks, but also for standard link prediction and node classification, and our results can also be used as baselines for these tasks. Looking forward, we hope that this work motivates further research proposing rich, high-quality, knowledge base benchmarks,  allowing the joint study of fact prediction across all arities based on KB structure and feature information.  %

\section*{Broader Impact}
Processing and completing knowledge graphs constitutes one of the main challenges towards producing high-quality, reliable automated inference methods, which benefit a multitude of application areas. This work proposes a means of completing these knowledge graphs in a unified fashion, and bases itself on an interpretable and explainable framework. This work also enables the use of node features, and thus introduces additional learning ability that benefits class and link prediction, and helps generalize the study of knowledge graphs towards settings more in line with real-world data.  

\begin{ack}
This work was supported by the Alan Turing Institute under the UK EPSRC grant EP/N510129/1, the AXA Research Fund, and by the EPSRC grants EP/R013667/1, EP/L012138/1, and EP/M025268/1. Ralph Abboud is funded by the Oxford-DeepMind Graduate Scholarship and the Alun Hughes Graduate Scholarship. Experiments for this work were conducted on servers provided by the Advanced Research Computing (ARC) cluster administered by the University of Oxford.
\end{ack}

\bibliographystyle{plain}
\bibliography{main.bib}

\appendix
\appendix
\section{Model Instantiations}
In this section, we present model-specific \framework instantiations of TransE, RotatE, and BoxE, the embedding models used in the main paper. For every model, we first introduce the base model, then provide the full definition of its extension under the \framework framework.

\subsection{MLP-TransE}
\begin{figure}
    \centering
    \includegraphics[width=\textwidth]{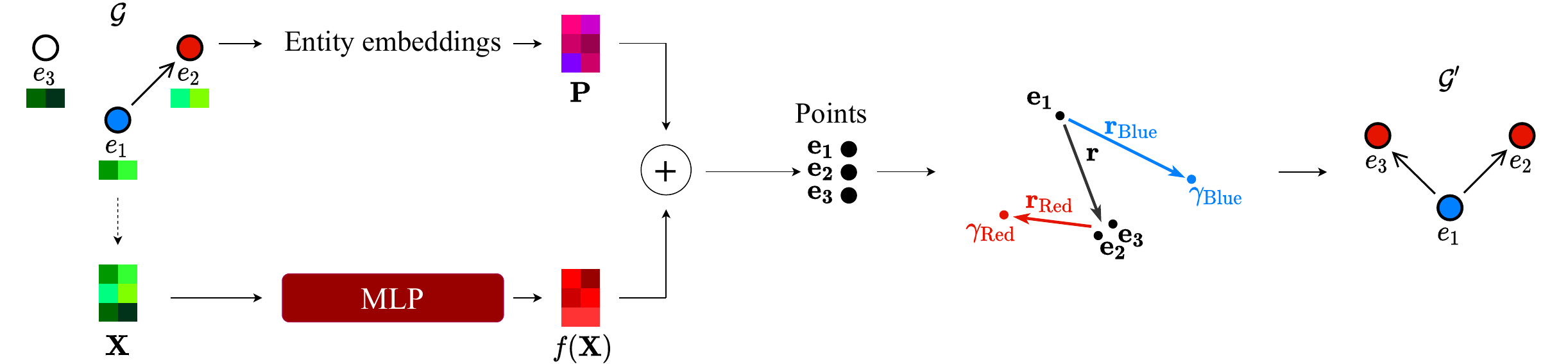}
    \caption{Given a KG $\mathcal{G}=\{Blue(e_1), Red(e_2),r(e_1, e_2)\}$ over entities $e_1,e_2,e_3$, MLP-TransE processes node features $\Xbf$ using an MLP $f$, and separately instantiates point embeddings $\Pbf$ for every entity. Then, these embeddings are summed with their corresponding MLP outputs to obtain representations in the TransE space, which, with the help of binarization, then predicts the graph $\mathcal{G'}= \mathcal{G} \cup \{Red(e_3),r(e_1, e_3)\}$.}
    \label{fig:MLPTransE}
\end{figure}

TransE \cite{TransE-NIPS13} is a seminal translational model for knowledge graph completion. It represents entities as points in a low-dimensional real space, and (binary) relations as translations in this space, such that entities linked by a relation are separated by its corresponding translation vector. More formally, in TransE, every entity $e_i \in \Ebf$ is represented by a vector $\bm{e_i} \in \Rbb^d$, and a relation $r$ is represented by a translation vector $\bm{r} \in \Rbb^d$. Using these representations, the likelihood score of a binary fact $r(e_h, e_t)$ is given by:
\begin{align*}
    \score(r(e_h, e_t)) &= \left\Vert \bm{e_h} + \bm{r} - \bm{e_t}  \right\Vert_2.
\end{align*}
TransE is a simple model for link prediction, and captures several inference patterns, such as anti-symmetry and composition. However, it has limitations handling one-to-many relations, e.g., $\mathsf{friends(Alice, Bob), friends(Alice, Charlie)}$ and capturing symmetry, e.g., the relation $\mathsf{knows}$. 

To supplement TransE, we apply an MLP on node features and combine with existing entity embeddings using a weighted summation. More formally:
\begin{align*}
    \bm{e_i'} = \lambda \bm{e_i} + f(\bm{x_i}),
\end{align*}
where $f: \Rbb^k \to \Rbb^d$ is an MLP, and $\lambda \in \Rbb^{+}$ is a scalar hyperparameter. Notice that this combination does not improve model expressiveness. For instance, any configuration that TransE cannot capture will clearly not be captured when node features are identical. 

As TransE does not support classes, we use the binarization described in the main paper. Note that the limited expressiveness of TransE also affects node classification predictions. For instance, capturing the facts $C(a), r(a, b), C(b), r(b, c)$, via their binarized counterparts $r_c(a, \gamma_c), r(a, b), r_c(b, \gamma_c), r(b, c)$ falsely implies $C(c)$, as the two binarized $r_c$ facts imply $\bm{e_a} = \bm{e_b}$, and thus that $\bm{r} = 0$ as $r(a, b)$ holds. Hence, $a=b=c$ and $C(c)$ holds.

An illustration for MLP-TransE is shown in \Cref{fig:MLPTransE}.

\subsection{MLP-RotatE}
\begin{figure}
    \centering
    \includegraphics[width=\textwidth]{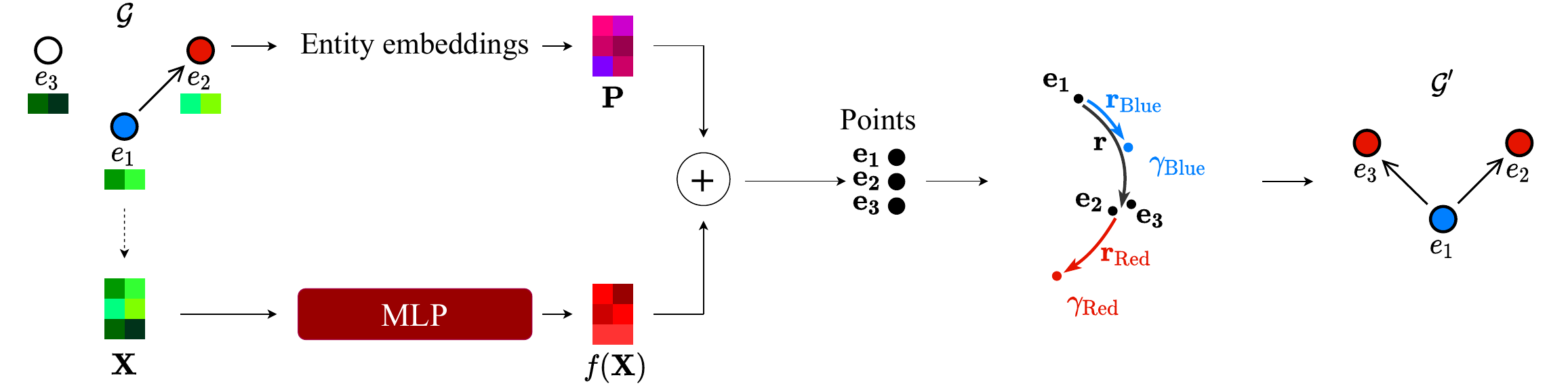}
    \caption{Given a KG $\mathcal{G}=\{Blue(e_1), Red(e_2),r(e_1, e_2)\}$ over entities $e_1,e_2,e_3$, MLP-RotatE processes node features $\Xbf$ using an MLP $f$, and separately instantiates point embeddings $\Pbf $for every entity. Then, these embeddings are summed with their corresponding MLP outputs to obtain representations in the RotatE space, which, with the help of binarization, then predicts the graph $\mathcal{G'}= \mathcal{G} \cup \{Red(e_3),r(e_1, e_3)\}$.}
    \label{fig:MLPRotatE}
\end{figure}
Analogously to TransE, RotatE \cite{RotatE-ICLR19} is a translational model. However, unlike TransE, it uses rotation, rather than translation, to represent relational structure. To this end,  it represents entities as points in a low-dimensional \emph{complex} space, and (binary) relations as \emph{norm-preserving rotations} in this space, such that entities linked by a relation are separated  by its corresponding rotation. More formally, every entity $e_i \in \Ebf$ is represented by a vector $\bm{e_i} \in \Cbb^d$, and a relation $r$ is represented by rotation vector $\bm{r} \in \Cbb^d$, where, for every dimension x in $\{0, ..., d-1\}$, $|\bm{r}[x]|=1$. Hence, the likelihood score of a binary fact $r(e_h, e_t)$ in RotatE is given by:
\begin{align*}
    \score(r(e_h, e_t)) &= \left\Vert \bm{e_h} \circ \bm{r} - \bm{e_t}  \right\Vert_2, 
\end{align*}
where $\circ$ denotes the element-wise rotation operation.

Compared with TransE, RotatE additionally supports relation symmetry, as this can be modelled using rotations of 180 degrees. However, it also suffers from the remaining limitations of TransE, namely handling one-to-many relations.

Analogously to TransE, we apply an MLP on node features and combine with existing entity embeddings using a weighted summation. More formally:
\begin{align*}
    \bm{e_i'} = \lambda \bm{e_i} + f(\bm{x_i}),
\end{align*}
where $f: \Rbb^k \to \Cbb^d$ is an MLP, and $\lambda \in \Rbb^{+}$ is a scalar hyperparameter. Note that producing outputs in a complex space using an MLP can be done by a reduction to $\Rbb^{2d}$. As with TransE, adding an MLP does not improve model expressiveness, and binarization is used to support classes. Finally, the limited expressiveness of RotatE also affects node classification, with the same example provided for MLP-TransE also holding for MLP-RotatE. 

An illustration for MLP-RotatE is shown in \Cref{fig:MLPRotatE}.
\subsection{MLP-BoxE}

\label{ssec:boxe}
BoxE \cite{BoxE-NeurIPS20} is a spatio-translational knowledge base completion model that can jointly predict facts for \emph{arbitrary} arity relations, and thus captures classes and binary relations as a special case.
For ease of presentation, we introduce BoxE on KGs, where we only have classes and binary relations.
BoxE operates in a low-dimensional space, and uses a \emph{region-based} semantics to determine the correctness of facts, based on the positions of their underlying \emph{entity}, \emph{class}, and \emph{relation} representations. More concretely, BoxE represents classes and binary relations using axis-aligned \emph{hyper-rectangles}, i.e., boxes, and these boxes fundamentally define regions in which said class and relation facts are true.

In BoxE, every entity $e_i \in \Ebf$ is represented by two vectors, a \emph{base position} vector $\bm{e_i} \in \mathbb{R}^d,$ and a \emph{translational bump} vector $\bm{b_i} \in \mathbb{R}^d$ , which translates all the entities co-occurring in a fact with $e_i$. 
The \emph{final embedding} of $e_i$ in a unary fact $c(e_i)$ is simply its position embedding $\bm{e_i}$, i.e., no translations occur. For a binary fact $r(e_h,e_t)$, the final embeddings of head entity $e_h$ (resp., tail entity $e_t$) is obtained by translating its position with the bump vector of the tail entity (resp.,  head entity):
\begin{align*}
\bm{e_h^{r(e_{h}, e_{t})}} = \bm{e_h} + \bm{b_{t}},  \quad \quad
\bm{e_t^{r(e_{h}, e_{t})}} = \bm{e_t} + \bm{b_{h}}.
\end{align*}

BoxE represents every class $c \in \Cbf$ by a single $d$-dimensional box $\bm{c}$, and every relation $r$ by a $d$-dimensional head box $\bm{r^{h}}$ and a $d$-dimensional tail box $\bm{r^{t}}$. Based on these representations, a plausibility score for unary and binary facts is computed respectively as:
\begin{align*}
    \score(c(e)) &= \left\Vert \dist(\bm{e}, \bm{c})  \right\Vert_x,  \\
    \score(r(e_h, e_t)) &= \left\Vert \dist(\bm{e_h^{r(e_h, e_t)}}, \bm{r^{h}})  \right\Vert_x + \left\Vert \dist(\bm{e_t^{r(e_h, e_t)}}, \bm{r^{t}})  \right\Vert_x,
\end{align*}
where $\dist$ intuitively computes the distance between a point and a box, and $x$ indicates the L-$x$ norm. 

\begin{wrapfigure}[14]{r}{0.48\textwidth}
    \centering
    \includegraphics[width=\linewidth]{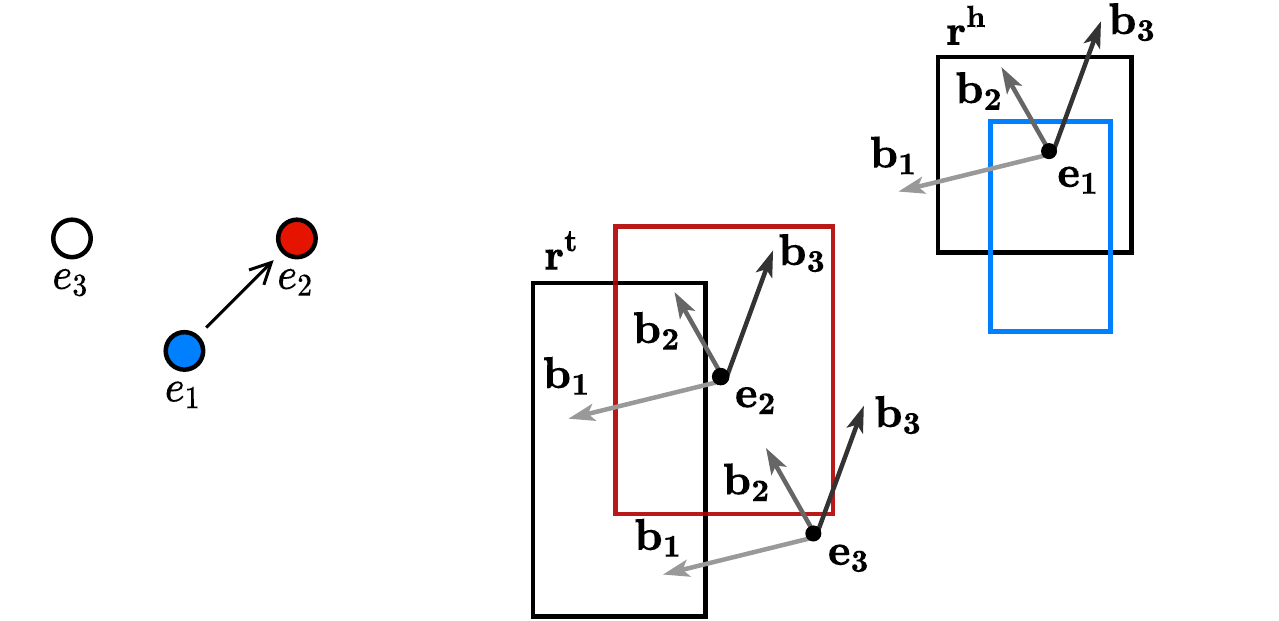}
    \caption{A knowledge graph (left) and an equivalent BoxE configuration (right).} %
    \label{fig:boxe_warm}
\end{wrapfigure}

Let us illustrate these notions on a KG with two unary facts $Blue(e_1)$, $Red(e_2)$ and one binary fact $r(e_1,e_2)$. This KG is shown in \Cref{fig:boxe_warm} along with its BoxE embedding.
Classes are depicted by blue and red node fills in the KG, and by blue and red boxes respectively in the BoxE space. Similarly, the binary relation $r$ is depicted as a black arrow in the KG, and by two black boxes $\bm{r^h}$ and $\bm{r^t}$ in the BoxE space. Moreover, position and bump vectors for entities $e_1, e_2,$ and $e_3$ are represented by labeled points and grayscale arrows respectively, and bump vectors are applied to all points to illustrate all possible pairwise interactions between entities. In this example, $\bm{e_1}$ is inside the blue box and $\bm{e_2}$ is inside the red box, whereas $\bm{e_3}$ is in neither, thereby matching all KG unary facts. Moreover, for binary relations, only $r(e_1, e_2)$ holds, as $\bm{e_1} + \bm{b_2}$ lies in $\bm{r^{h}}$, and $\bm{e_2} + \bm{b_1}$ lies in $\bm{r^{t}}$, but this is not true for any other entity combination.

\begin{figure}
    \centering
    \includegraphics[width=\textwidth]{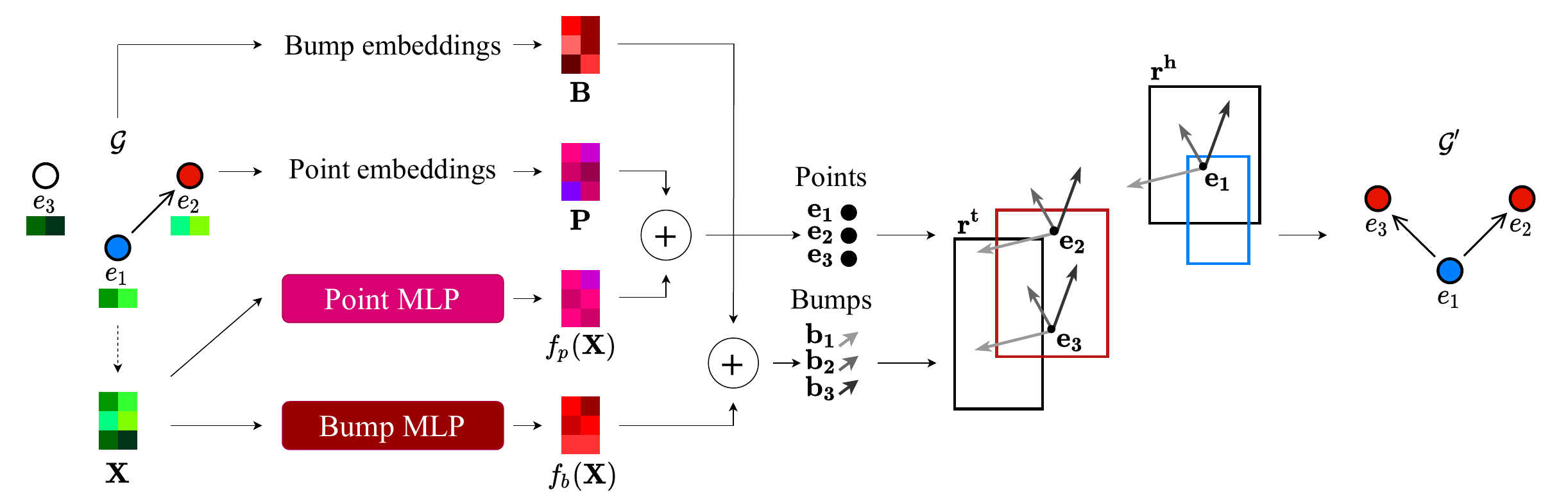}
    \caption{Given a KG $\mathcal{G}=\{Blue(e_1), Red(e_2),r(e_1, e_2)\}$ over entities $e_1,e_2,e_3$, MLP-BoxE processes node features $\Xbf$ using two (point and bump) MLPs $f_p$ and $f_b$, and separately instantiates point ($\Pbf$) and bump ($\Bbf$) embeddings for every entity. Then, these embeddings are summed with their corresponding MLP outputs, yielding representations in the BoxE space which are used to predict the graph $\mathcal{G'}= \mathcal{G} \cup \{Red(e_3),r(e_1, e_3)\}$.}
    \label{fig:overview_model}
\end{figure}

Given a KG, MLP-BoxE processes node features $\Xbf$ in a row-wise fashion using \emph{two distinct} MLPs $f_p$ and $f_b$.%
 More concretely, given $\bm{x}_i$, the $i^\text{th}$ row vector in $\Xbf$ and the feature vector for $e_i$, both MLPs return $f_p(\bm{x}_i), f_b(\bm{x}_i) \in \Rbb^d$ respectively, and these vectors are the row vectors for $f_p(\Xbf), f_b(\Xbf) \in \Rbb^{|\Ebf|\times d}$. In parallel, MLP-BoxE instantiates point and bump embedding matrices $\Pbf, \Bbf \in \Rbb^{|\Ebf|\times d}$, such that every entity has point and bump embedding vectors $\bm{e_i}$ and $\bm{b_i}$ respectively, corresponding to the $i^{\text{th}}$ row vectors of both matrices. Finally, MLP-BoxE produces position and bump representations for all entities as:
\begin{align*}
\bm{e'_i} = \lambda\bm{e_i} + f_p(\bm{x_{i}}),  \quad \quad
\bm{b'_i} = \lambda\bm{b_i} + f_b(\bm{x_{i}}),
\end{align*}
where $\lambda$ is a scalar hyper-parameter.
At this point, MLP-BoxE defines boxes for relations and classes, and performs scoring analogously to standard BoxE and \Cref{ssec:boxe}. %
The overall structure of MLP-BoxE is shown in \Cref{fig:overview_model}.

Intuitively, MLP-BoxE processes node features and uses them to inform its representations, while also respecting the incompleteness assumption and not structurally enforcing neighborhoods as with GNNs. In particular, MLPs provide just enough power to process features, but then delegate all relational processing and feature combination to the BoxE  space. This enables new edges and feature combinations to implicitly be learned where needed, maintains relational inductive bias, and creates a feedback loop between the embedding space and the MLP during training. %
As BoxE is fully expressive, MLP-BoxE is also fully expressive given a sufficient dimensionality $d$, which is a corollary to Theorem 5.1 of \cite{BoxE-NeurIPS20}.

\begin{corollary}
\label{col:exp}
Consider a relational vocabulary, consisting of finite sets $\Ebf$ of entities, $\Cbf$ of classes, and $\Rbf$ of relations. Let $\Smc$ be the set of all facts over this vocabulary, where every entity has a matching feature from a feature matrix $\Xbf \in \Rbb^{|\Ebf|\times k}$. Then, for any set $\Tmc \subseteq \Smc$ of true and any disjoint set $\Fmc \subseteq \Smc$ of false facts, there exists a MLP-BoxE model $M$ with dimensionality $d = |\mathbf{E}||\mathbf{R}| + |\mathbf{C}|$ that maps all facts of $\Tmc$ to true and all facts of $\Fmc$ to false.
\end{corollary}

\begin{proof}
It is known that BoxE can represent a knowledge graph over $|\Ebf|=n$ entities and $|\mathbf{R}|$ binary relations using a dimensionality $n|\mathbf{R}|$ \cite{BoxE-NeurIPS20}. This result can be extended to include unary classes simply by considering unary classes as special binary relations with a fixed head (or tail), yielding a dimensionality bound of $n(|\mathbf{R}|+|\Cbf|)$. However, we propose an extended construction in this corollary to capture unary classes with a lower dimensionality, namely $n|\mathbf{R}|+|\Cbf|$. This construction begins from the $n|\mathbf{R}|$-dimensional BoxE configuration capturing all binary facts, resulting from Theorem 5.1 in the BoxE paper \cite{BoxE-NeurIPS20}, and extends it with new dimensions so as to capture $\Cbf$ and all class memberships.

More specifically, let $\bm{e_m}$ and $\bm{b_m}$ denote position and bump embeddings for every entity $e_m$ in BoxE, and, for every relation $r_j$, let ${\bm{r}_j}^h$ and ${\bm{r}_j}^t$ denote their head and tail boxes, and let $\bm{l}_{j}^{h}, \bm{l}_{j}^{t}$, and $\bm{u}_j^{h}, \bm{u}_j^{t}$ indicate the lower and upper corners of both boxes respectively. Finally, we denote the $k^\text{th}$ dimension value of a vector $\bm{v}$ by $v(k)$.

Given a BoxE configuration capturing all binary facts in a KG, following BoxE Theorem 5.1, we now additionally define the class boxes and enforce class memberships as follows:
\begin{enumerate}
    \item For every class $c_i\in \mathbf{C}$, define a box $\bm{c}_i$, such that for $k \in \{1, ..., n|\mathbf{R}|\}$, its lower corner is set as $\bm{l}_i(k) = \min_{m} \bm{e}_m(k) - \epsilon$, $\epsilon \in \mathbb{R}, \epsilon > 0$,  whereas its upper corner is defined as $\bm{u}_i^{(1)}(k) = \max_{m} \bm{e}_m(k) + \epsilon$. Intuitively, this defines the class boxes in the first $n|\mathbf{R}|$ dimensions to fit all points, and thus to hold true in these dimensions for all entities. 
    \item For $i \in \{1, ..., |\mathbf{C}|\}$, define additional dimensions, such that $\bm{l}_i(n|\mathbf{R}| + i) = -1$ and $\bm{u}_i(n|\mathbf{R}| + i) = 1$. This defines a new dimension per class, such that the corresponding class box in said dimension spans the interval $[-1, 1]$.
    \item For $m \in \{1, ..., n\}$, and $i \in \{1, ..., |\mathbf{C}|\}$,
    \begin{enumerate}
        \item If $c_i(e_m) \in \Tmc$, set $\bm{e}_m(n|\mathbf{R}| + i) = 0$, and thus place it inside $\bm{c}_i$.
        \item Otherwise, if $c_i(e_m) \in \Fmc$, set $\bm{e}_m(n|\mathbf{R}| + i) = 2$, outside $\bm{c}_i$.
    \end{enumerate}
    This step sets entity positions for entities belonging to a class $c_i$ inside its box, in its dedicated dimension, and vice-versa. 
    
    \item Bump vectors are inconsequential for fitting unary classes, and thus are set to 0 in all added dimensions, i.e., $\forall i \in \{1, ..., |\mathbf{C}|\}, \bm{b}_m(n|\mathbf{R}| + i) = 0$.
    
    \item Finally, we now ensure all the addition of new dimensions does not render previously true binary facts false, by setting the relation boxes to include all possible class values in the added $|\Cbf|$ dimensions. More concretely, for $i \in \{1, ..., |\mathbf{C}|\}$,  for $j \in \{1, ..., |\mathbf{R}|\}$: 
    \begin{enumerate}
        \item $\bm{l}_j^{h}(n|\mathbf{R}| + i) = -3$, $ \bm{l}_j^{t}(n|\mathbf{R}| + i) = -3$,
        \item $\bm{u}_j^{h}(n|\mathbf{R}| + i) = 3$, $ \bm{u}_j^{t}(n|\mathbf{R}| + i) = 3$.
    \end{enumerate} 
    Hence, all relation boxes cover the interval $[-3, 3]$ in the added dimensions, and thus include all entity representations in those dimensions by construction. Hence, all previously correct binary facts remain true with the added dimensions.
\end{enumerate}
Therefore, BoxE admits a parametrization $\mathcal{W}$ with $d = n|\mathbf{R}| + |\mathbf{C}|$ that can represent all possible configurations of a knowledge graph over classes $\mathbf{C}$ and relations $\mathbf{R}$. This naturally extends to MLP-BoxE. In particular, given BoxE configuration $\mathcal{W}$, and given MLPs $f_p$ and $f_b$, we can exactly reconstruct $\mathcal{W}$ using MLP-BoxE, simply by setting its embedding vectors for every entity $e_m$ as:
\begin{align*}
\bar{\bm{e}}_m &= \bm{e'}_{m, \mathcal{W}} - f_p(\mathbf{x}_m), \text{and} \\
\bar{\bm{b}}_m &= \bm{b'}_{m, \mathcal{W}} - f_b(\mathbf{x}_m), 
\end{align*}
where the $\mathcal{W}$ subscript indicates that this vector originates from parametrization $\mathcal{W}$, and disambiguates from the corresponding vector in MLP-BoxE.
\end{proof}

\section{\datasetname Class and Relation Statistics}  %
 Fundamentally, \datasetname proposes 24 balanced and mutually exclusive classes, each containing exactly 700 entities, as well as 29 relations with highly different semantics, and with varying prominence in the data. Classes and relations in \datasetname are not semantically connected, and thus a class cannot be predicted simply by detecting a given edge type, and vice-versa. More concretely, knowing the alma mater of an entity does not inform about the existence of any relation, e.g. $\mathsf{actor}$, and no relation gives away a specific alma mater. Hence, \datasetname avoids the interdependence between classes and relations found in current citation graph benchmarks \cite{OGB-NeurIPS2020}, e.g., the two nodes for a $\mathsf{cites}$ edge are necessarily papers. 
 
Moreover, \datasetname proposes highly distinct relations which do not have redundancies between them. In particular, no relations trivially map to one another, e.g., inverse relations, hierarchy, which makes the link prediction task challenging. Therefore, \datasetname is designed to serve as a strong benchmark for link prediction, in keeping with recently proposed refinements to standard link prediction datasets \cite{ConvE-AAAI18, toutanova2015observed}.

The class and relation statistics for our newly-proposed \datasetname dataset can be found in \Cref{tab:wiki}.
\begin{table}[h]
\caption{The classes and relations in \datasetname.}
\label{tab:wiki}
\centering
\begin{tabular}{lrlr} 
    \toprule
    \multicolumn{2}{c}{\textbf{Classes}} & \multicolumn{2}{c}{\textbf{Relations}}\\
    \midrule
    Class Name & Appearances & Relation Name & Appearances \\
    \cmidrule{1-2}
    \cmidrule{3-4}
    Brown University & 700 & about & 21 \\
    Columbia University & 700 & actor & 15752 \\
    Cornell University & 700 &  affiliation & 27 \\
    Harvard University & 700 &  author & 3360 \\
    Massachusetts Institute of Technology & 700 & award & 8031 \\
    Michigan State University & 700 & birthPlace & 14086\\
    Northwestern University & 700 & character & 24\\
    New York University & 700 & children & 2028 \\
    Ohio State University & 700 & competitor & 75\\
    Princeton University & 700 & composer  & 155\\
    Syracuse University & 700 &  contributor  & 11\\
    United States Military Academy & 700 & creator & 873 \\   
    United States Naval Academy & 700 & deathPlace  & 7143 \\
    University of California, Berkeley & 700 & director & 3761\\
    University of California, Los Angeles & 700 & editor & 159\\
    University of Chicago & 700 & founder & 482 \\
    University of Michigan & 700 & gender & 18 \\
    University of Pennsylvania & 700 & hasOccupation & 20572\\
    University of Texas at Austin & 700 & homeLocation & 1187 \\
    University of Toronto & 700 & knowsLanguage  & 4847\\
    University of Virginia & 700 & lyricist  & 111\\
    University of Washington & 700 & memberOf & 12337\\
    University of Wisconsin-Madison & 700 & musicBy & 1794 \\
    Yale University & 700 &  nationality & 16382\\
     & & parent & 2021\\
     & & producer & 4267\\
     & & publisher & 51\\
     & & spouse & 2242 \\
     & & worksFor  & 19\\
    \bottomrule
    \end{tabular}
\end{table}

\section{Additional Experiments}
In this section, we report additional experiments that complement our findings in the main paper. First, we experiment with MLP-BoxE, the best-performing node classification model in our main experiments, on single-relational citation graph benchmarks, and observe that it performs strongly even in this restricted setting. Second, we experiment with poorer bag-of-word features on \datasetname with all \framework models, and observe that these models respond better to the decline in feature quality. Finally, we  experiment with BoxE on YAGO39K to compare with TransC \cite{LiHW18}, and compare it against TuckER \cite{TuckER} on \datasetname-80\%. We also discuss the performance of bilinear models on our dataset. 

\subsection{Experiments on Standard Citation Graph Benchmarks}

\begin{table}[t]
    \centering
    \caption{Node classification results (accuracy) for MLP-BoxE and competing models on public splits of standard benchmarks (Test set). Results reported for other models are the best published.}
    \label{tab:citation_results}
    \begin{tabular}{lcccc}
        \toprule
         Model &  CiteSeer & Cora & PubMed & OGBN-arXiv \\
         \cmidrule(r){2-2}
         \cmidrule(r){3-3}
         \cmidrule(r){4-4}
         \cmidrule(r){5-5}
         MLP \cite{VelickovicICLR18,OGB-NeurIPS2020} & 46.5 & 55.1 & 71.4 & 55.5 \\
         GCN \cite{VelickovicICLR18,Kipf16,OGB-NeurIPS2020} & 70.9 & 81.5 & 79.0 & \textbf{71.7}\\
         GAT \cite{VelickovicICLR18} & \textbf{72.5} & \textbf{83.0} & 79.0 & \_ \\
         MLP-BoxE & 70.2 & 78.4 & \textbf{81.4} & 69.4\\
         \bottomrule
    \end{tabular}
\end{table}

\begin{table}[t]
    \centering
    \caption{Hyper-parameters for MLP-BoxE on the node classification benchmarks.}
    \label{tab:citation_hp}
    \begin{tabular}{lccc}
        \toprule
         Dataset & Margin & Dimensionality  & Dropout \\
         \cmidrule(r){2-4}
         CiteSeer & 2 & 20 & 0.8 \\
         Cora & 2 & 20 & 0 \\
         PubMed & 2 & 20 & 0.5 \\
         OGBN-arXiv & 3 & 256 & 0 \\
         \bottomrule
    \end{tabular}
\end{table}

To evaluate the modeling ability of \framework models beyond our main study, we train MLP-BoxE, the best-performing node classification model among \framework models, on three standard citation network benchmarks, namely Citeseer, Cora, and PubMed \cite{SenNBGGE08}, as well as the recently introduced OGBN-arXiv \cite{OGB-NeurIPS2020} benchmark. 

For Citeseer, Cora, and PubMed, we observed that MLP-BoxE suffered from  overfitting with a two-layer MLP, and thus only used a single hidden layer of size 1000. Furthermore, we used dropout on the MLPs, and tuned this using grid search over the range [0,1] in increments of 0.1, and this yielded substantial performance improvement on PubMed and Citeseer. We tuned embedding dimensionality, and found that, surprisingly, MLP-BoxE performed best and in the most stable fashion with only 20 dimensions, and we believe this is due to the sparsity and low label rate of these benchmarks. For OGBN-arXiv, we used the standard dimensionality of 256 used in the literature, a two-layer MLP, as in the main experiments, and train using negative sampling loss. 

In all experiments, we used a learning rate of 0.001, negative sampling loss, and use 10 negative samples per positive fact. Furthermore, we modified negative sampling for unary facts as described in the main paper. The final hyper-parameters used are shown in \Cref{tab:citation_hp}, and the results are shown in \Cref{tab:citation_results}.

Across all datasets, MLP-BoxE performs strongly, and remains competitive with GNNs, even achieving state-of-the-art performance on PubMed. This is very encouraging, as these benchmarks are single-relational and, with the exception of OGBN-arXiv, highly sparse in terms of node labels. Therefore, MLP-BoxE performs competitively even in this very specialised setting. %

\subsection{Experiments with Bag-of-Words Features on \datasetname}

To evaluate the impact of feature quality on different node classification models, we perform an ablation study on the features of \datasetname. In particular, we replace the original GloVe mean vector features with a binarized bag-of-words feature vector, consisting of the 100 most frequent words across the corpus of all entity Wikipedia descriptions, and re-train all models on the new dataset analogously to the original setup. As in the main paper, we train \framework models with both cross-entropy and negative sampling loss, and report the latter as it offered the better results in this setup. In fact, the optimal hyper-parameters for MLP-TransE and MLP-RotatE exactly coincide with those for negative sampling loss in the entity classification setting (cf. \Cref{tab:NS}), whereas the optimal margin for MLP-BoxE is 3. The results for this ablation study are shown in \Cref{tab:bow}.

\begin{wraptable}{r}{7cm}
    \centering
    \caption{Results with 100-dimensional bag-of-words features on the full \datasetname dataset.}
    \label{tab:bow}
    \begin{tabular}{lc}
        \toprule
        Model & Accuracy (\%) \\
        \midrule
        MLP & 19.0 \\
        rGCN & 33.1\\
        GAT & 30.9 \\
        \midrule
        MLP-TransE(NS) & \third{34.3}\\
        MLP-RotatE(NS) & \first{35.1}\\
        MLP-BoxE(NS) & \second{34.9}\\  %
        \bottomrule
    \end{tabular}
\end{wraptable}

In this setup, the performance of all models drops severely, as expected. However, GAT drops the most substantially, even falling behind rGCN. On the other hand,  \framework models now comfortably outperform GAT, despite the relative completeness of relational data in this setting. Across the board, all models comfortably beat the MLP baseline, and do so by a higher margin relative to the original experiment, which suggests that all models exploit edge information to compensate for lost features.

From these results, we see that GAT heavily relies on feature quality, and suffers greatly when features are poor, or unavailable, as in entity classification. rGCN also relies on features, but responds better to the loss of feature quality, as it uses relational structure more effectively. Finally, \framework models, owing to their relational inductive bias, more effectively exploit relational information, and now gains an advantage against GNNs, as these models can no longer make up the performance gap through feature processing.

\subsection{Additional Comparative Experiments}

We additionally conduct experiments with BoxE and TransC \cite{LvHLL18}, two models with class represeentation ability,  on YAGO39K, and observed that the former achieves \emph{very significant} improvement on the latter. These results motivated our selection of BoxE for the main paper. 
We ran BoxE with the same dimensionality $d=100$ as TransC, and use d negative sampling loss with margin 6, and uniform negative sampling with 100 negative samples per positive. Results are reported in \Cref{tab:boxe}.

\begin{table}[h!]
    \centering
    \caption{BoxE results versus TransC on YAGO39K.}
    \label{tab:boxe}
    \begin{tabular}{lcc}
        \toprule
        Model & MRR & H@10\\
        \midrule
        TransC \cite{LvHLL18} & 0.421 & 0.698 \\
        BoxE & \textbf{0.542} & \textbf{0.782}\\
        \bottomrule
    \end{tabular}
\end{table}

Furthermore, we ran 128-dimensional TuckER \cite{TuckER} over \datasetname-80\% without classes and features, as this is the only setting where it naturally applies. There, we observed that our base \framework models achieve superior performance. Results for this experiment are reported in \Cref{tab:boxe_tucker}.

\begin{table}[h!]
    \centering
    \caption{TuckER, TransE, RotatE, and BoxE results on \datasetname-80\% (no classes and features).}
    \label{tab:boxe_tucker}
    \begin{tabular}{lccc}
        \toprule
        Model & MR & MRR & H@10\\
        \midrule
        TuckER \cite{TuckER} & 4681 & 0.248 & 0.333 \\
        TransE(NS) & 1843 & 0.265 & \textbf{0.376}\\
        RotatE(NS) & \textbf{1806} & \textbf{0.285} & 0.375\\
        BoxE(CE) & 2298 & \textbf{0.285} & 0.366\\
        
        \bottomrule
    \end{tabular}
\end{table}

Finally, we evaluated both TuckER and ComplEx when conducting our experiments for the main paper, and observed that these models performed substantially worse than their translational counterparts. For instance, ComplEx would only achieve a best node classification accuracy of 35.5\%, whereas TuckER would achieve at most 28.5\% on entity classification, despite extensive tuning efforts. We then found that this performance is primarily due to the high mean rank both models achieve, which negatively impacts node classification performance, as explained in the main paper. More precisely, both models have substantially worse MR than the three used \framework models, and this implies that link prediction quality, though potentially better in specific cases, is worse overall, and thus predicted links are much noisier, leading to overall worse performance in our setting. 

\section{Details of the Main Experiments}

\begin{table}[t]
    \centering
    \caption{Hyper-parameter settings for \framework models with \emph{negative sampling loss}. Here, $\alpha$ denotes the learning rate used by the Adam optimizer.}
    \label{tab:NS}
    \begin{tabular}{lccccccccc}
        \toprule
         \multirow{3}{*}{Model} & \multicolumn{3}{c}{\datasetname-80\%}  & \multicolumn{3}{c}{\datasetname-90\%}  & \multicolumn{3}{c}{\datasetname}\\
         \cmidrule{2-4}
         \cmidrule{5-7}
         \cmidrule{8-10}
         & $\alpha$ & Margin & \makecell{Batch\\Size} & $\alpha$ & Margin & \makecell{Batch\\Size}& $\alpha$ & Margin & \makecell{Batch\\Size}\\
        \midrule
        MLP-TransE & $10^{-4}$ & 15 & 1024 & $10^{-4}$ & 15 & 1024 & $10^{-4}$ & 15 & 1024\\
        MLP-RotatE & $10^{-4}$ & 21 & 1024 & $10^{-4}$ & 21 & 1024 & $10^{-4}$ & 24 & 1024 \\
        MLP-BoxE & $10^{-3}$ & 9 & 512 & $10^{-3}$ & 6 & 512 & $10^{-3}$ & 6 & 512 \\
        \midrule
        TransE & $10^{-3}$ & 9 & 1024 & $10^{-3}$ & 9 & 1024 & $10^{-3}$ & 9 & 1024\\
        RotatE & $10^{-3}$ & 9 & 1024 & $10^{-3}$ & 9 & 1024 & $10^{-3}$ & 9 & 1024\\
        BoxE & $10^{-3}$ & 4 & 512 & $10^{-3}$ & 5 & 512 & $10^{-3}$ & 5 & 512 \\
        \bottomrule
    \end{tabular}
\end{table}

\begin{table}[t]
    \centering
    \caption{Hyper-parameter settings for \framework models with \emph{cross-entropy loss}. Here, $\alpha$ denotes the learning rate used by the Adam optimizer.}
    \label{tab:CE}
    \begin{tabular}{lcccccc}
        \toprule
         \multirow{3}{*}{Model} & \multicolumn{2}{c}{\datasetname-80\%}  & \multicolumn{2}{c}{\datasetname-90\%}  & \multicolumn{2}{c}{\datasetname}\\
         \cmidrule{2-3}
         \cmidrule{4-5}
         \cmidrule{6-7}
         & $\alpha$ & \makecell{Batch\\Size} & $\alpha$ & \makecell{Batch\\Size}& $\alpha$ & \makecell{Batch\\Size}\\
        \midrule
        MLP-TransE & $10^{-4}$ & 128 & $10^{-4}$ & 128 & $10^{-4}$ & 128\\
        MLP-RotatE & $10^{-4}$ & 1024 & $10^{-4}$ & 128 & $10^{-4}$ & 128 \\
        MLP-BoxE & $10^{-3}$ & 512 & $10^{-3}$ & 512 & $10^{-3}$ & 128 \\
        \midrule
        TransE & $10^{-3}$ & 1024 & $10^{-3}$ & 1024 & $10^{-3}$ & 1024\\
        RotatE & $10^{-3}$ & 1024 & $10^{-3}$ & 1024 & $10^{-3}$ & 1024\\
        BoxE & $10^{-3}$ & 512 & $10^{-3}$ & 512 & $10^{-3}$ & 512 \\
        \bottomrule
    \end{tabular}
\end{table}

In our experiments, all \framework models were trained on a Haswell CPU node with 12 cores, 64 GB RAM, and a V100 GPU. The results reported are (i) the mean peak validation accuracy across 5 runs for node classification, and (ii) the mean of MR/MRR/Hits@10 results on all dropped edges across 5 runs, using the best average MRR-Hits@10 as the validation metric for link prediction. In our experiments, we observed that the error bars in terms of accuracy (0.2\%) and MRR (0.1\%) are small and consistent across models, and thus did not report them for better visibility. 

When training the models, we set both all MLPs to have two hidden layers of size 1000, each using the ReLU activation layer. Moreover, we conducted training using the Adam optimizer \cite{Kingma-ICLR2014}, 100 negative samples per positive fact, and a learning rate $\alpha$ chosen from the set $\{10^{-4}, 10^{-3}\}$. Negative sampling was conducted as described in the main paper, by negatively sampling over classes for class facts, to exploit the mutual exclusion between target classes. Furthermore, we experimented with the values 0.5 and 1 for $\lambda$, the embedding scale, and found that 0.5 achieved better results across all experiments and provided better regularization for the models.

We additionally experimented with two different loss functions: negative sampling loss (NS) \cite{RotatE-ICLR19} (with uniform negative sampling), and cross-entropy loss (CE), the standard loss formulation for node classification, and reported results for both functions in the main paper. Finally, we tuned batch size with values from the interval \{128, 256, 512, 1024\} and, for negative sampling loss, also tuned loss margin, from the range \{1, 2, 3, 4, 5, 6, 9, 12, 15, 18, 21, 24, 27, 30\}. The hyper-parameters corresponding to our results for negative sampling loss are reported in \Cref{tab:NS}, and those corresponding to our results for cross-entropy are reported in \Cref{tab:CE}. We note that, in our experiments, the optimal hyper-parameters for node classification and link prediction with/without features coincide, despite being tuned independently, and this further highlights the interdependence between both tasks. Furthermore, the optimal hyperparameters for link prediction for all configurations without classes are identical to those for the corresponding configurations with classes, i.e., the same hyperparameters achieve best performance for a given configuration with and without classes, all else equal.

\end{document}